\documentclass[journal]{IEEEtran}
\usepackage{bm}
\usepackage{booktabs}
\usepackage[switch,pagewise,columnwise]{lineno}

\usepackage{cite}

\ifCLASSINFOpdf
  \usepackage[pdftex]{graphicx}
  \graphicspath{{IEEEtran/figures/}}
  \DeclareGraphicsExtensions{.pdf,.jpeg,.png}
  \usepackage[justification=centering]{caption}

\else
\fi
\usepackage{amsmath}
\usepackage{algorithmic}

\usepackage{array}
\begin{document}
\linenumbers
%
\title{Dynamic Balancing of Humanoid Robot Walker3 with Proprioceptive Actuation: 
Systematic Design of Algorithm, Software and Hardware}
%
%
%

\author{Yan Xie$^{1}$, Jiajun Wang$^{1}$, Hao Dong$^{1}$, Xiaoyu Ren$^{1}$, Liqun Huang$^{1}$, Mingguo Zhao$^{2}$
\thanks{$^{1}$Yan Xie, Jiajun Wang, Hao Dong, Xiaoyu Ren and Liqun Huang are with the Beijing Research Institute of UBTECH Robotics, Beijing, China. Email: xieyan\_buaa@163.com.}
\thanks{$^{2}$Mingguo Zhao is with the Department of Automation, Tsinghua University and Beijing Innovation Center for Future Chips, Tsinghua University, Beijing, China. Email: mgzhao@tsinghua.edu.cn.}}

\maketitle

\begin{abstract}
Dynamic balancing under uncertain disturbances is important for a humanoid robot, which requires a good capability of coordinating the entire body redundancy to execute multi tasks. Whole-body control (WBC) based on hierarchical optimization has been generally accepted and utilized in torque-controlled robots. A good hierarchy is the prerequisite for WBC and can be predefined according to prior knowledge. However, the real-time computation would be problematic in the physical applications considering the computational complexity of WBC. For robots with proprioceptive actuation, the joint friction in gear reducer would also degrade the torque tracking performance. In our paper, a reasonable hierarchy of tasks and constraints is first customized for robot dynamic balancing. Then a real-time WBC is implemented via a computationally efficient WBC software. Such a method is solved on a modular master control system UBTMaster characterized by the real-time communication and powerful computing capability. After the joint friction being well covered by the model identification, extensive experiments on various balancing scenarios are conducted on a humanoid Walker3 with proprioceptive actuation. The robot shows an outstanding balance performance even under external impulses as well as the two feet of the robot suffering the inclination and shift disturbances independently. The results demonstrate that with the strict hierarchy, real-time computation and joint friction being handled carefully, the robot with proprioceptive actuation can manage the dynamic physical interactions with the unstructured environments well.
\end{abstract}


%
\IEEEpeerreviewmaketitle

\section{Introduction}
%
%
%
%
\IEEEPARstart{H}{}umanoid robots have been widely studied for decades for a better mobility, a higher dexterity and a stronger intelligence. Making humanoids work in human-tailored environments or interact with humans, however, still faces enormous problems. As one of the most fundamental skills for legged robots, balance maintenance is of primary importance for a humanoid. The balance control approaches based on simplified models has been well studied in early works. The robot is simplified as a single rigid body mounted on the top of an inverted pendulum with the regulation of center of mass (CoM) as control objective \cite{2016li}. Further, a simplified three-link planar model with more information is proposed in \cite{2013nenchev}. These approaches utilizing the ankle or hip joints have been proved to be practical on real robots, however, suffer from an unnatural behavior due to failing to account for all joints motion.

To fully exploit the capability of redundant joints, whole-body control (WBC) has been proposed and gained a growing interest within the robotics community over the past twenty years. Up to now, the existing WBCs can be categorised into: (1) null space projection based WBC (NSP-WBC), (2) weight quadratic program based WBC (wQP-WBC) and (3) hierarchical quadratic program based WBC (hQP-WBC). The NSP-WBC can implements the hierarchy of tasks via null space projections. Such an idea originates from the redundancy control of robot manipulators \cite{1987nakamura} and is first extended to the walking of HRP-2 robot in \cite{2003kajita}. However, such a method is limited as it can not take the inequality constraints into account. For torque-controlled robots, dynamic constraints such as joint torque limits are highly necessary for robots' safety. The wQP-WBC can handle this problem by formulating the tasks and constraints as a quadratic optimization problem. It can find the optimal solution that minimizes the task errors while satisfying the constraints. This method has been applied on Atlas robot to execute multi tasks during the DARPA Robotics Challenge \cite{2014dai}\cite{2015feng}\cite{2016Koolen}. However, the strict task hierarchy can not be guaranteed in this method. Only soft hierarchy can be realized by tuning the task weight. This limitation becomes a concern when the tasks conflict with each other.

The hQP-WBC has been devised to combine the task hierarchy and inequality constraints together \cite{2010de} \cite{2011kanoun} \cite{2014escande}. The basic idea is constructing each layer as a quadratic optimization problem and solving them in a sequence where the lower priority QPs can not disturb the higher priority QPs. As a result, the inequality constraints in QP stack gradually as the priority decreases, thus leading to a time consuming problem. \cite{2016herzog} proposed an efficient way by introducing null space projection to reduce the computational cost and implemented it on a torque-controlled robot Sarcos with a 1KHz control loop. Similarly, \cite{2016bellicoso} implemented this method on a quadruped robot ANYmal showing a natural adaption to the terrain while walking. In short, the hQP-WBC has been a genetic whole-body generation tool for torque-controlled robots.

For achieving highly dynamic motions, proprioceptive actuation has been proposed and implemented in legged robots \cite{2012seok}. The torque is controlled directly by the motor's current with the torque loss in the gear reducer being well covered. Thanks to its high bandwidth and high torque density, the MIT Cheetah robot demonstrated the dynamic contact interactions in its high-speed locomotion \cite{2017wensing}. However, the joint friction could be negligible as the gear reduction ratio becomes larger, thus requiring an off-line model identification process.

Besides the high bandwidth characteristic of actuators, the overall performance of the robot is highly depends on the control frequency. \cite{2016kim} has noticed a phenomenon in the torque-controlled robot Mercury that increasing the control frequency from 1KHz to 1.5KHz will provide a larger posture and foot position control bandwidth. This puts forward a higher demand for the real-time computation of WBC. Moreover, a faster WBC is also preferred as other time-consuming techniques such as model predictive control are embedded into the control framework. 

This paper mainly focus on the the systematic integration of algorithm, software and hardware in the dynamic balancing of a humanoid robot with proprioceptive actuation. We first customize a reasonable hierarchy of tasks and constraints for adapting to disturbances. Then a reduced whole-body control is implemented and solved in real-time by a computationally efficient WBC software. A modular master control system UBTMaster is also designed for providing real-time communication and powerful computing capability in hardware side. Next, a model identification process is conducted on the robot to cover the joint friction and model inaccuracy issues. Finally, extensive experiments on various balancing scenarios are implemented on a robot Walker3 with proprioceptive actuation. The balance performance under various kinds of disturbance are discussed.

\begin{figure}[htp]
    \centering
    \includegraphics[width=3.5in]{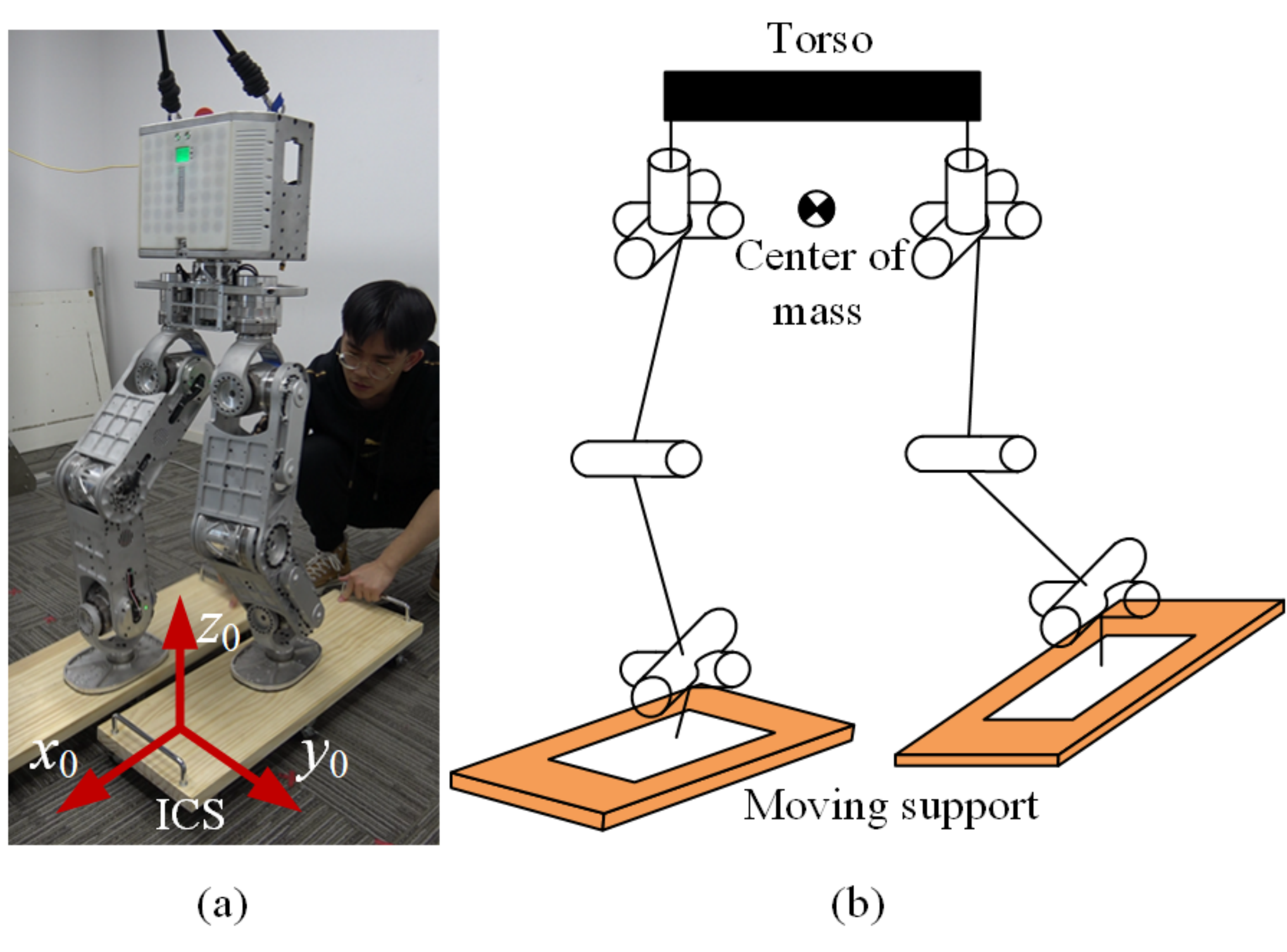}
    \caption{Walker3 humanoid robot with 12 actuated DoFs (a), and its kinematic model (b).}
    \label{fig_robot_model}
\end{figure}

Walker3 is a humanoid robot consisting of two legs and a torso as shown in Fig. \ref{fig_robot_model}. The robot is 1.6m tall and weights 43Kg. Inertial measurement unit (IMU) is mounted on the torso for the state estimation. Two 6-axis force sensors are installed on the soles to measure the center of pressure (CoP) in each foot. Each leg has 6 electrical motors in series. The motors use gear reducer to enlarge the output torque and the gear reduction ratio is up to 100. The motors are real-time controlled using the EtherCAT communication protocol.

\section{Control Approach}
To adapt to the uncertain disturbances, an online task planner is proposed to make adjustments on the task trajectory. The basic idea of this planner is to ensure the foot ZMP resides in the safe region as much as possible. Then the desired trajectory is tracked by a reduced whole-body controller coupled with a hierarchical optimization solver. The hierarchy of tasks and constraints is divided into four layers according to their priorities. Each layer is solved by a quadratic optimization and the strict hierarchy among layers is guaranteed by the null space projection. After solving a sequence of QPs, the whole-body controller outputs the optimized joint torques.

\begin{figure*}[htp]
    \centering
    \includegraphics[width=7.0in]{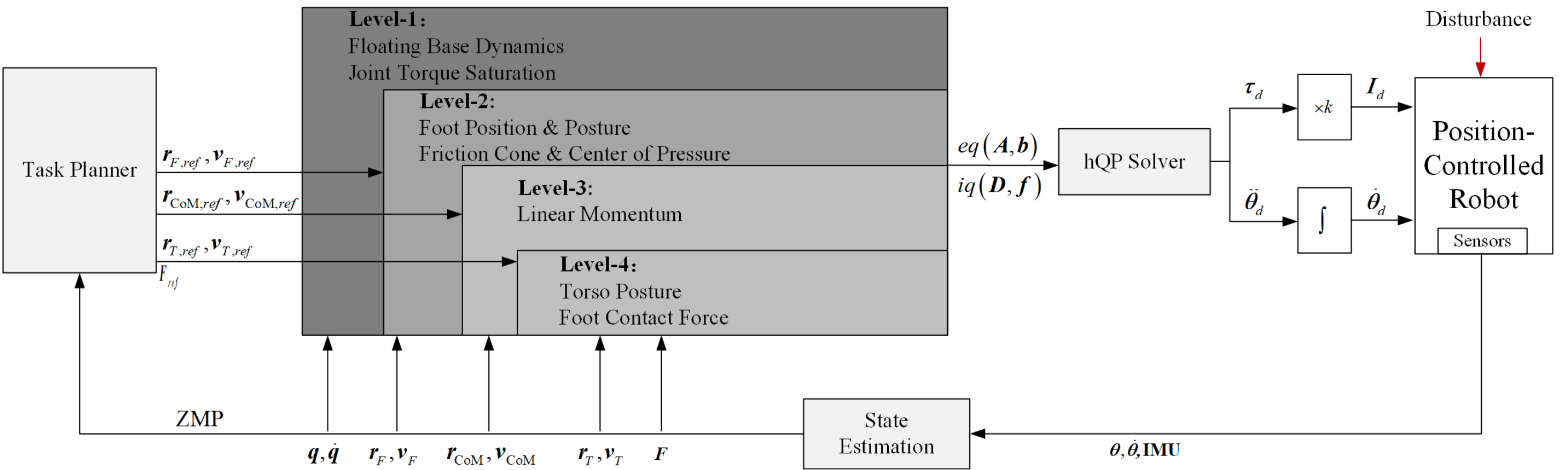}
    \caption{Overview of the control architecture}
    \label{fig_control_architecture}
\end{figure*}

For a robot with proprioceptive actuation, joint torques are turned into joint current commands. The desired joint velocity commands obtained through numerical integration of desired joint accelerations are also considered here to improve the performance of the joint-level control. The whole control architecture is illustrated in Fig. \ref{fig_control_architecture}. The state of the robot is estimated by the kinematics solver with the sensory data of joints and IMU.

\section{Tasks and constraints in dynamic balancing}

\subsection{Task planner}
\label{Sec_control_approach_ZMP_based_planner}

Facing dynamic environments, robot needs to tune the desired motion of the task properly. The task planner presented here can make real-time adjustments on the task trajectory according to the ZMP state. 

Different from the method in \cite{2016herzog}, we have no extra sensors to get information about the moving support but the force sensors to detect the contact between foot and support. When a robot is stably resting on a moving support, the ZMP of each foot must reside inside its support polygon. Therefore, the desired motion of the foot can be set using the following rule: if the measured ZMP is inside the safe region of the support polygon, we adjust the desired position, posture and velocity of the foot to its current state ${r_{F,ref}} = {r_F} $ and ${v_{F,ref}} = {v_F}$.

In addition, the desired motion of CoM should be adjusted along with the foot desired motion. The desired horizontal position is located in the middle of two feet $ {r_{{\rm{CoM}},ref}}\left( {x,y} \right){\rm{ = }}\left( {{r_{LF,ref}}\left( {x,y} \right) + {r_{RF,ref}}\left( {x,y} \right)} \right)/2 $ and the desired vertical position is set to $ {r_{{\rm{CoM}},ref}}\left( z \right) = \left( {{r_{LF,ref}}\left( z \right) + {r_{RF,ref}}\left( z \right)} \right)/2 + {\rm{C}} $. $\rm C $ is a constant value depending on the robot stand pose. The desired velocity is set to the average velocity of two feet $ {v_{{\rm{CoM}},ref}}{\rm{ = }}\left( {{v_{LF,ref}} + {v_{RF,ref}}} \right)/2 $. In our planner, the desired motion of torso and foot contact force remain unchanged throughout the balance control. The desired motion of torso is set to $ {r_{T,ref}}{\rm{ = }}0, {v_{T,ref}}{\rm{ = }}0 $ and the desired vertical contact force is set to $ {F_{LF,ref}}\left( z \right){\rm{ = }}{F_{RF,ref}}\left( z \right){\rm{ = }}mg/2 $, where $m$ is the total mass of the robot.

\subsection{Tasks and constraints hierarchy}
\label{Sec_control_approach_tasks_and_constraints_hierarchy}

When multi tasks have to be performed simultaneously, how to handle the conflicts among these objectives is crucial. Prioritized hierarchy strategy has been widely adopted in redundant robots. Motion solvers will accomplish the lower priority tasks as well as possible under the prerequisite that the higher priority tasks is implemented first. For example, balance is always considered a top-priority task for a humanoid robot. The robot tends to sacrifice its posture under disturbance to ensure the feet is fully in contact with the ground. Likewise, physical constraints concerning the humanoid robot safety should be put in the highest priority. Table. \ref{table_task_constraint_hierarchy} specifies the hierarchy of tasks and constraints. 

\begin{table*}[htbp]
    \label{table_task_constraint_hierarchy}
    \centering
    \caption{Tasks and constraints hierarchy}
    \renewcommand\arraystretch{1.5}
    \begin{tabular}{@{}lllll@{}}
    \toprule
    Level & Task                                                                       & Task Dimensions & Constraint                                                                 & Constraint Dimensions \\ \midrule
    1     & Floating base dynamics                                                         & 6               & Joint torque saturation                                                    & 12                    \\
    2     & Foot position and posture                                                   & 12              & \begin{tabular}[c]{@{}l@{}}Center of pressure\\ Friction cone\end{tabular} & 18                    \\
    3     & Linear momentum                                                            & 3               &                                                             &                      \\
    4     & \begin{tabular}[c]{@{}l@{}}Torso posture\\ Foot contact force\end{tabular} & 15              &                                                                            &                       \\ \bottomrule
    \end{tabular}
    \label{table_task_constraint_hierarchy}
\end{table*}

\subsubsection{Floating base dynamics}
Humanoid robots, typical floating-based systems, are a particular case of underactuated systems due to the partial actuation when they make contact with the environment. The configuration of a humanoid robot is represented by generalized coordinates $ q = {\left[ {\begin{array}{*{20}{c}}
{q_f^{\rm{T}}}&{q_a^T}
\end{array}} \right]^{\rm{T}}} $, where $ {q_f} $ represents the position and orientation of the robot free-floating body and $ {q_a} $ represents the $ {n} $ actuated joints of the robot. When the robot is in contact with the environment, the dynamic equation of the system can be fully described by

\begin{equation}
    {S_f}M\ddot q + {S_f}C + {S_f}G = {S_f}{J^{\rm{T}}}{F}
\end{equation}

where $ {M} $ is the generalized inertia matrix, $ {C} $ is the nonlinear vector including Coriolis and centrifugal forces, $ {G} $ is the gravity vector. $ {F} $ is the contact force vector. $ {J} $ is the jacobian matrix of the contact point. $ {S_f} = \left[ {\begin{array}{*{20}{c}}
I&0
\end{array}} \right] $ is a matrix selecting the free-floating joints. Thus, actuated torque vector $ {\tau} $ is eliminated from the dynamic equation. Instead, choosing a different selection matrix $ {S_a} = \left[ {\begin{array}{*{20}{c}}
0&I
\end{array}} \right] $ will derive a linear function between $ {\tau} $ and $ {\ddot q} $, $ {F} $. Due to such linear dependence, the full body dynamics of 
$ {n+6} $ dimensions are simplified as the floating base dynamics of 6 dimensions. The adoption of floating base dynamics is crucial to reduce the optimization time and will help to implement the 1KHz control loop.

As the highest priority task, dynamics equation is the essential description for a physical multi-rigid-body system. Once the equation holds, the movements of the system are physically feasible. The second priority task following the dynamics equation is the foot position and posture control. A good task control performance will guarantee a good contact between foot and support, which is a premise for the contact force. Linear momentum control, the third priority task, has been proved to be essential for a good balance by regulating the state of CoM \cite{2012lee}. In the lowest priority task, we prefer to have the torso posture control and foot contact force control on the same level. Given that we have only 30 optimization variables (including 18 for $ {\ddot q} $ and 12 for $ {F} $) while dynamic equation, foot position and posture task and linear momentum task lock 6, 12 and 3 variables separately, only 9 free variables are left for the torso posture and foot contact force control.

\subsubsection{Operational space tasks}
Operational space tasks such as foot position and posture control, linear momentum control and torso posture control can be phrased as
\begin{equation}
    {J}\ddot q + {\dot J}\dot q = {a}
\end{equation}
where $ {J} $ is the jacobian matrix of a specific task, $ a $ is the task desired acceleration which can be determined by a feedforward and feedback control law. The $ {\dot J}\dot q $ term is related to the robot state. Specially, the jacobian matrix of the linear momentum task is also named as centroidal momentum matrix and can be calculated using an efficient O($ n $) algorithm based on the generalized inertia matrix $ {M} $ \cite{2016wensing}.

\subsubsection{Robot safety constraints}
Given the physical limitations of the robot, several safety issues have to be appropriately concerned. The joint torque saturation constraint $ {\tau _{\min }} \le \tau  \le {\tau _{\max }} $ is especially important for generating control commands that are valid on a robot.

A stable contact between foot and support is an essential precondition for generating a 6-dimensional contact force vector, which further means the foot is not allowed to tilt or slide relative to the support. The stable contact can be ensured from two aspects. One is the center of pressure constraint. The center of pressure at each foot must no exceed the boundary of the foot's support polygon. The other is the friction cone constraint, which requires the foot contact force must stay inside the friction cones. The cones are approximated as pyramids here so that the constraint can be expressed as linear inequality.

\section{Real-time WBC}

\subsection{Reduced hierarichal whole-body control}
\label{Sec_control_approach_whole_body_control}

The tasks can be formulated as equalities and constraints can be formulated as inequalities. Therefore, the tasks and constraints in the same level can be stacked vertically into the form

\begin{equation}
    \label{eq:task_and_constraint}
    \left\{ {\begin{array}{*{20}{c}}
    {{{{A}}_i}{{x}} - {{{b}}_i} = 0}\\
    {{{{D}}_i}{{x}} - {{{f}}_i} \le 0}
    \end{array}} \right.
\end{equation}

where $ {{{A}}_i} $ is the $ i $th task matrix, $ {{{b}}_i} $ is the $ i $th task reference vector, $ {{{D}}_i} $ is the $ i $th constraint matrix, $ {{{f}}_i} $ is the $ i $th constraint boundary vector, $ {{x}} = {\left[ {\begin{array}{*{20}{c}}
{{{{{\ddot q}}}^{\rm{T}}}}&{{{{F}}^{\rm{T}}}}
\end{array}} \right]^{\rm{T}}} $ is the optimal variables. The goal of this level is to find $ {{{\ddot q}}} $ and $ {{F}} $ that satisfies these objectives as well as possible. The solution under such a linear inequality constraint can be solved through quadratic optimization. The tasks and constraints in different levels need to be optimized in a strict prioritized order. Solving level $ p $ yields an optimal solution $ {{x}}_p^* $. In order to ensure the strict prioritization of tasks, the solution of level $ p+1 $ can be found in the null space of all higher priority tasks $ {{N}}_p^{} = {{N}}_{p - 1}^{}\left( {{{I}} - {{\hat A}}_p^\# {{{{\hat A}}}_p}} \right) $. $ {{N}}_{p - 1}^{} $ is the null space of all tasks from level 1 to $ p-1 $. $ \left( {{{I}} - {{\hat A}}_p^\# {{{{\hat A}}}_p}} \right) $ is the null space of task in level $ p $. $ {{\hat A}_p} = {A_p}{N_{p - 1}} $ describes the task matrix of level $p$ projected into the null space of all higher priority tasks. The solution of level $ p+1 $ can be expressed as ${{{x}}_{p + 1}}{ = }{{x}}_p^* + {{{N}}_p}{{{u}}_{p + 1}}$, where ${{{u}}_{p + 1}}$ is an arbitrary vector lying in the row space of ${{{N}}_p}$. Substituting ${{{x}}_{p + 1}}$ into the QP problem in level $ p+1 $ yields

\begin{equation}
    \begin{array}{l}
    \mathop {\min .}\limits_{{{{u}}_{p + 1}}} \;\;\;\;\;\;\;{\left\| {{{{A}}_{p + 1}}\left( {{{x}}_p^* + {{{N}}_p}{{{u}}_{p + 1}}} \right) - {{{b}}_{p + 1}}} \right\|^2}\\
    \;\;\;{\rm{s}}.{\rm{t}}.\;\;\;\;\;\;\;\;\;\;\;\;\;{{{D}}_{p + 1}}\left( {{{x}}_p^* + {{{N}}_p}{{{u}}_{p + 1}}} \right) - {{{f}}_{p + 1}} \le 0\\
    \;\;\;\;\;\;\;\;\;\;\;\;\;\;\;\;\;\;\;\;\;\;{{{D}}_p}\left( {{{x}}_p^* + {{{N}}_p}{{{u}}_{p + 1}}} \right) - {{{f}}_p} \le 0\\
    \;\;\;\;\;\;\;\;\;\;\;\;\;\;\;\;\;\;\;\;\;\; \vdots \\
    \;\;\;\;\;\;\;\;\;\;\;\;\;\;\;\;\;\;\;\;\;\;{{{D}}_1}\left( {{{x}}_1^* + {{{N}}_p}{{{u}}_{p + 1}}} \right) - {{{f}}_1} \le 0
    \end{array}
\end{equation}

All the higher priority constraints are stacked into the optimization to ensure the strict prioritization of constraints. Using this recursive algorithm, QP in each level is solved sequentially according to its priority.

Different from the general formulation in \cite{2016herzog}, the slack variables are omitted in the optimization problem. The slack variables are originally introduced to turn the hard constraint into a soft one. In our case, however, we notice that the optimized slack variables are always zero which means that the solver can find the optimal result without violating the hard constraints. Therefore, the slacks are excluded from the optimization variables to reduce the computational complexity. Such a method is not general yet proved to be computationally efficient in the physical applications.

\begin{figure*}[htp]
    \centering
    \includegraphics[width=7.0in]{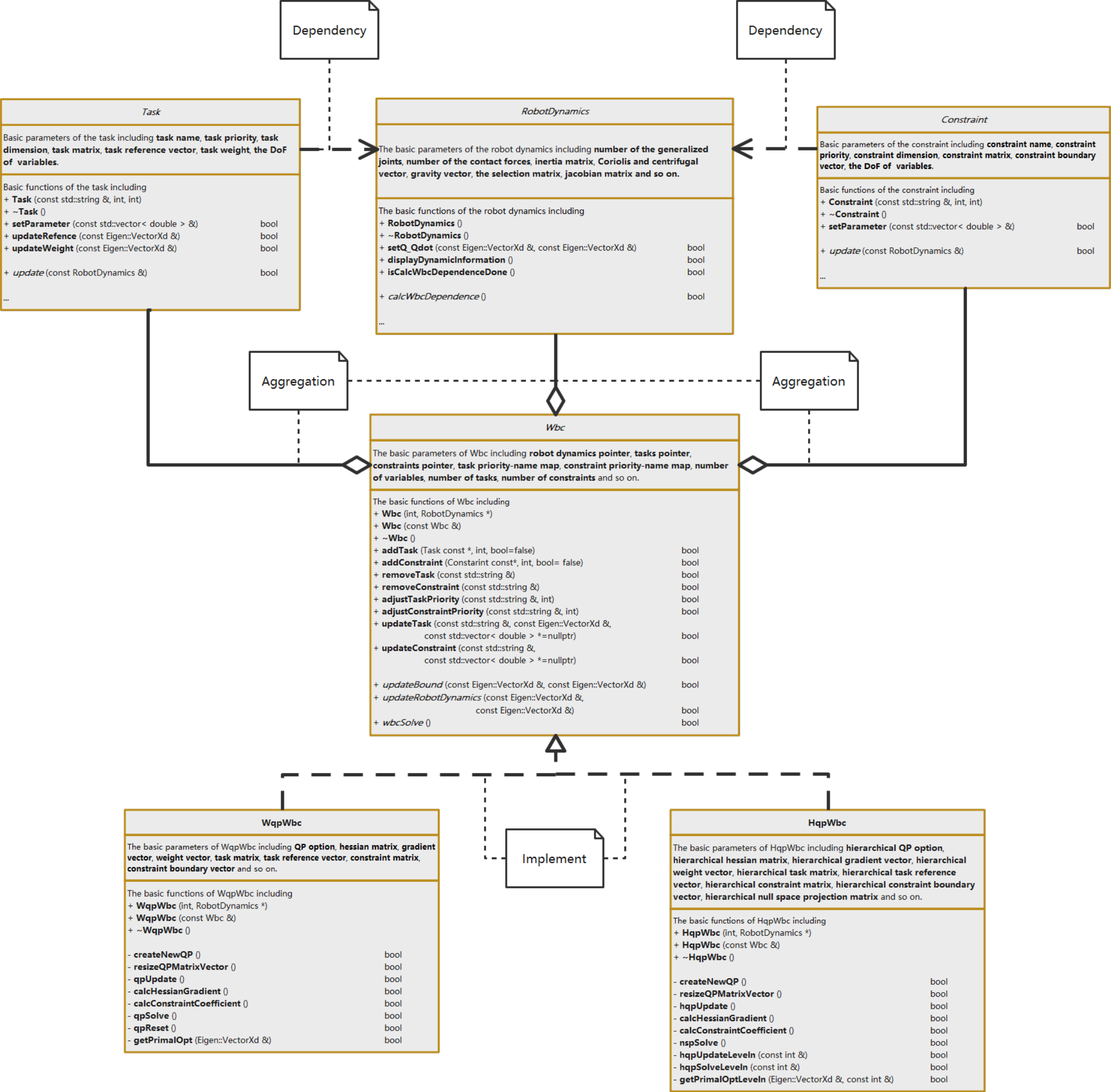}
    \caption{The architecture of WBC software}
    \label{fig_wbc_architecture}
\end{figure*}

To implement the above algorithm in an effective way, a WBC software is developed based on C++. Fig. \ref{fig_wbc_architecture} depicts the architecture of the WBC software. The software contains four basic classes: RobotDynamics, Task, Constraint and Wbc. These classes provide the basic interfaces for user development and the derived classes about Walker3 is developed in this software. The following part will describe these classes in detail.

The RobotDynamics class contains the member variables related to the kinematics and dynamics of a robot, such as the number of the generalized joints, number of the contact forces, inertia matrix, Coriolis and centrifugal vector, gravity vector, the selection matrix, jacobian matrix and so on. For Walker3 robot, it is implemented by a subclass named RobotDynamics\_Walker3. The model structure of Walker3 can be constructed in the subclass directly or loaded from URDF files. Calling the calcWbcDependence() function can get all the required kinematics and dynamics parameters. The open-source Rigid Body Dynamics Library (RBDL), a highly efficient C++ library that contains some essential rigid body dynamics algorithms \cite{2014featherstone}, is used here. 

The Task and Constraint classes are constructed according to Eq. (\ref{eq:task_and_constraint}) with their member variables including the task's or constraint's name, priority, dimension, matrix, reference vector (boundary vector) and the DoF of variables. Each task or constraint of Walker3 is implemented by a subclass, thus forming a task or constraint library. Calling the update(const RobotDynamics \&) function will update the member variables with calculated kinematics and dynamics parameters.

The Wbc class, the core of the software, contains the pointers of other three classes. It can manage the addition, deletion and adjustment operations of the tasks and constraints. As its implementation, two subclasses based on different algorithms is developed here. One is named WqpWbc which forms all the tasks and constraints as one quadratic optimization problem \cite{2015feng}. The other is named HqpWbc which implements the hierarchical quadratic optimization as mentioned before.

Benefiting from this software, lots of redundant codes are avoided to enhance the developing efficiency. Meanwhile, developers also build the dynamic model of a quadruped and its corresponding tasks' and constraints' library. Users can also build their own robots without rewriting the WBC solver code.

\subsection{High-performance master control system}

\begin{figure*}[htp]
    \centering
    \includegraphics[width=7.0in]{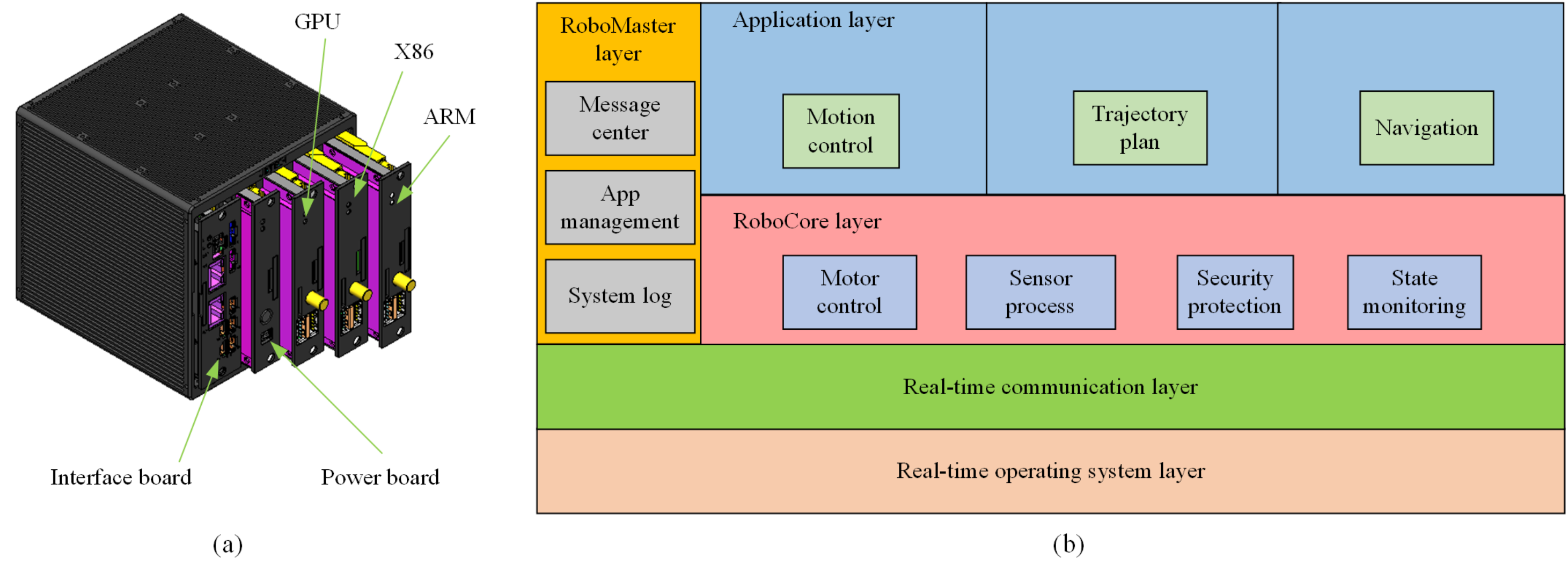}
    \caption{The modular master control system (a) and its software architecture (b)}
    \label{fig_master_control_system}
\end{figure*}

The WBC software is embedded on a modular master control system named UBTMaster as shown in Fig. \ref{fig_master_control_system}(a). It is designed with the characteristics of strong real-time computation, extensible computing capability and configurable interface. It can realize the different combination configurations of typical hardware platforms such as ARM, GPU, X86 and DSP, and allow the expansion of the computing capability in different scenarios. For the X86 basic edition, it can perform 100 GFLOPS per second with the Intel core i7-7600U.

The software architecture is illustrated in Fig. \ref{fig_master_control_system}(b). The real-time operating system based on the PREEMPT\_RT kernel is to serve real-time applications that process data as it comes in, typically without buffer delays. It ensures that the application's task must be done within the defined time constraints. In the real-time communication layer, we apply the high-speed, real-time bus communication protocol EtherCAT for short data update times (also called cycle times; $ {\le} $ 100 us) with low communication jitter (for precise synchronization purposes; $ {\le} $ 1 us). From these two aspects, the time jitter can be controlled at a microsecond level.

In the upper layer, the roboCore is running in a real-time process and used for isolating the applications from the hardware platform. Users can develop their own applications to meet the specific requirements.

\section{Proprioceptive actuation with large reduction ratio}

\subsection{Joint-level control}
\label{Sec_joint_level_control}
Given the inputs, WBC software will output the optimized joint torque. However, the lack of torque sensor in proprioceptive actuation do not allow the direct torque control. A common workaround for this problem is to utilize an admittance coupling to convert joint torque to joint velocity \cite{2012Dietrich}. Considering the bandwidth of admittance control will limit the torque tracking performance, here we utilize direct current control.

For proprioceptive actuation with small reduction ratio, the joint current can be approximated as a linear function of joint torque due to the negligible torque loss in the reducer. However, there is a significant amount of static friction in the reducer with a large reduction ratio. This stiction translates into joint torque stiction of up to 5 Nm. Consequently, joint friction torque compensation is essential. Besides, the joint velocity obtained by integrating the optimized joint acceleration can also be added to the joint current command as a kinematic compensation term to improve the joint impedance \cite{2016Koolen}. 

In our paper, the final control law for joint current is calculated as

\begin{equation}
    {i^{{\rm{cmd}}}}{\rm{ = }}{k_i}\left( {{\tau ^{{\rm{opt}}}}{\rm{ + }}{k_f}{\tau ^f} + {\tau ^{\dot q}}} \right)
\end{equation}

where $ {{\tau ^{{\rm{opt}}}}} $ is the optimized joint torque,  $ {{\tau ^f}} $ is the joint friction compensation torque, $ {{k_f}} $ is the corresponding friction compensation coefficient, $ {{\tau ^{\dot q}}} $ is the joint kinematic compensation torque. $ {{\tau ^{{{opt}}}}} $ ,$ {{\tau ^f}} $ and  $ {{\tau ^{\dot q}}} $ can be expressed as

\begin{equation}
    {\tau ^{opt}} {\rm{ = }}S_a^{}M{\ddot q^{opt}} + S_a^{}\left( {C + G} \right) - S_a^{}{J^{\rm{T}}}{F^{opt}}
\end{equation}

\begin{equation}
    {\tau ^f}{\rm{ = }}\left\{ {\begin{array}{*{20}{l}}
    {{F_c} + {F_v}\left( {\int {{{\ddot q}^{opt}}dt}  - {{\dot q}^*}} \right),{\kern 1pt} {\kern 1pt} {\kern 1pt} \int {{{\ddot q}^{opt}}} dt \ge {{\dot q}^*}}\\
    {\int {{{\ddot q}^{opt}}} dt\frac{{{F_c}}}{{{q^*}}},{\kern 1pt} {\kern 1pt} {\kern 1pt} {\kern 1pt} {\kern 1pt}  - {{\dot q}^*} \le \int {{{\ddot q}^{opt}}} dt \le {{\dot q}^*}}\\
    { - {F_c} + {F_v}\left( {\int {{{\ddot q}^{opt}}} dt + {{\dot q}^*}} \right),{\kern 1pt} {\kern 1pt} {\kern 1pt} \int {{{\ddot q}^{opt}}} dt \le  - {{\dot q}^*}}
    \end{array}} \right.
\end{equation}

\begin{equation}
    {\tau ^{\dot q}}{\rm{ = }}{k_{\dot q}}\left( {\int {{{\ddot q}^{{\rm{opt}}}}dt}  - \dot q} \right)
\end{equation}

where $ {{\tau ^{opt}}} $ is reorganized from the full dynamics. The joint friction compensation torque $ {\tau ^f} $ is modeled as two parts: coulomb and viscous friction. $ {F_c} $ is the coulomb friction and $ {F_v} $
is the viscous friction coefficient. $ {{\dot q}^*} $ is a user defined value to prevent a sudden jump in joint friction compensation torque as motor rotates reversely. $ {k_{\dot q}} $ is a gain acting on the difference between the desired joint velocity $ \int {{{\ddot q}^{{\rm{opt}}}}dt} $ and the measured joint velocity $ \dot q $.

\subsection{Model identification}
\label{sec_model_identification}
Model identification is an effective method to obtain a robot's dynamic parameters. Besides the concerned dynamic parameters such as links' mass, inertia and center of mass, joint friction model can also be incorporated into the linearlized dynamic equation \cite{2018Gaz}. Here, coulomb-viscous friction model is preferred due to its linear expression. 
    
The main process can be divided into three parts.
    
\subsubsection{Linearization of dynamic equation} 
Use recursive Newton-Euler equation to reorganize the joint torque $ {\tau} $ as a linear function of dynamic parameters $ {\pi} $  (including joint friction parameters), $ \tau  = Y\pi $. $ Y $ is the identification matrix and can be uniquely determined by joint motion $ q $, $ {\dot q} $ and $ {\ddot q} $.

\subsubsection{Optimal excitation trajectory}
The excitation trajectory is parameterized first by a finite Fourier series function and then optimized for the minimum of a user defined cost function \cite{2016siciliano} while satisfying the constraint conditions. The series and base frequency in Fourier series are set 5 and 0.1Hz separately. The condition number of Y matrix is closely relative to the mean square error of identification results and thus selected as the cost function.
    
\subsubsection{Dynamic parameters optimization}
Construct an optimization problem to find optimal dynamic parameters $ {\pi} $ which can minimize the error between the measured joint torque and the predicted joint torque by linear dynamic equation.

In addition to the motors lacking of torque sensors, a prior calibration of current-torque coefficient can help to get approximate joint torque through the measured joint current. 
    
\begin{figure*}[htp]
    \centering
    \includegraphics[width=7.0in]{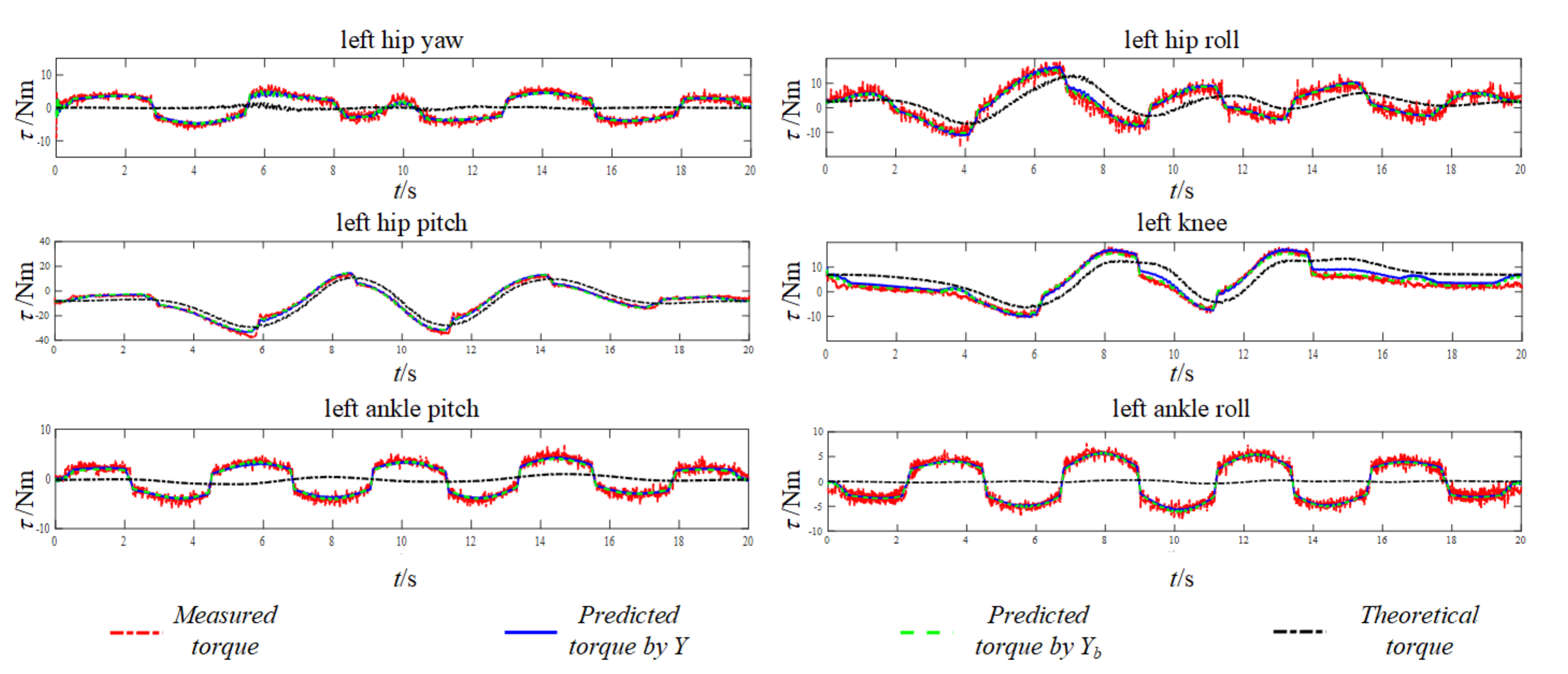}
    \caption{The comparison of joint torque in left leg}
    \label{fig_model_identification}
\end{figure*}

Fig. \ref{fig_model_identification} compares four sets of  joint torque in left leg. Red dotted line indicates the measured torque through joint current. Blue solid line indicates the predicted torque through identified dynamic parameters while green dotted line indicates the predicted torque through identified base dynamic parameters. Black dotted line indicates the theoretical torque calculated by parameters obtained from 3D model. Obviously, there is a large error between the measured and theoretical torque. The main reason is that the joint friction torque is not considered in the dynamics equation calculation. 
   
Not surprisingly, the error between the measured and predicted torque is very small. Meanwhile, several torque jumps are measured when the motor changes its rotating direction. Thanks to the coulomb friction term in our friction model, the predicted torque curve follows the measured one closely. This directly prove that the identified dynamic parameters can reflect the actual dynamic characteristics and can be used for modeling in the real control system. 
   
\section{Experimental validation}
The above mentioned control approach is experimentally evaluated on the Walker3 humanoid. The balance performance is evaluated in different scenarios: push recovery on the ground, balancing on a seesaw, push recovery on two moving skateboards. A summary of experimental videos is available in the supplementary materials.

\subsection{Push recovery on the ground}

\begin{figure*}[htp]
    \centering
    \includegraphics[width=6.0in]{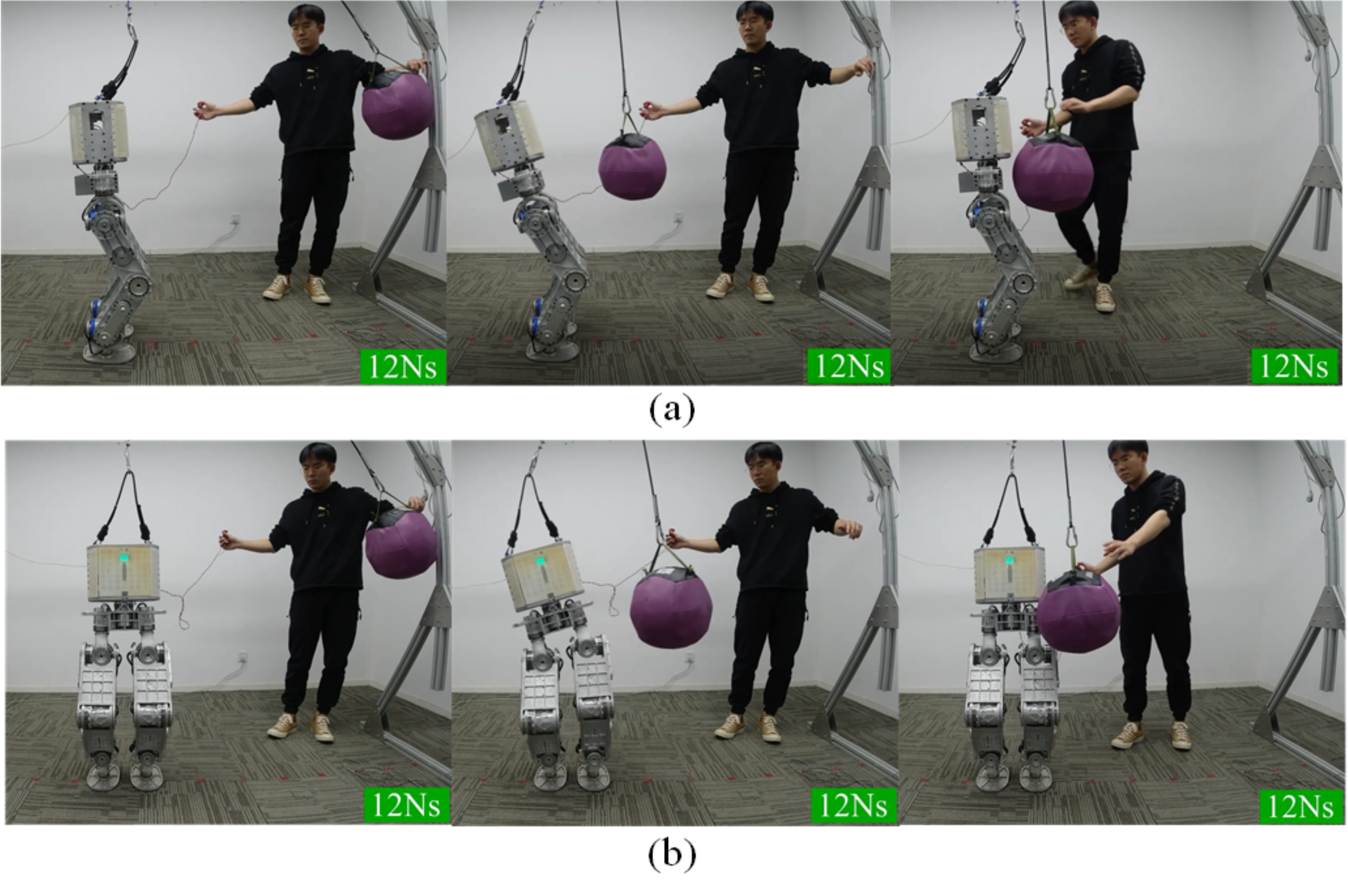}
    \caption{The balancing behavior in push recovery scenario along the X-axis (a), and Y-axis (b).}
    \label{fig_push_recovery}
\end{figure*}

The robot is subjected to impulses from X and Y directions while it is standing on the ground. A ball weighed 5Kg is used to generate impulses at the torso of the robot. The impulses can be quantified by the ball' momentum with the its known mass and velocity. Fig. \ref{fig_push_recovery} shows a series of snapshots when the robot is stroked along X-axis and Y-axis. Eight impulses along X-axis and seven impulses along Y-axis are exerted on the robot. 

For the push recovery along X-axis, the index and magnitude of impulses are listed in Tab. \ref{tab_impulses_magnitude}.

\begin{table}[htp]
    \centering
    \caption{The magnitude of impulses along X-axis}
    \label{tab_impulses_magnitude}
    \renewcommand\arraystretch{2}
    \begin{tabular}{lllll}
    \hline
    Index     & 1   & 2    & 3    & 4$\sim$8 \\ \hline
    Magnitude & 8Ns & 10Ns & 11Ns & 12Ns     \\ \hline
    \end{tabular}
\end{table}

\begin{figure}[htp]
    \centering
    \includegraphics[width=3.5in]{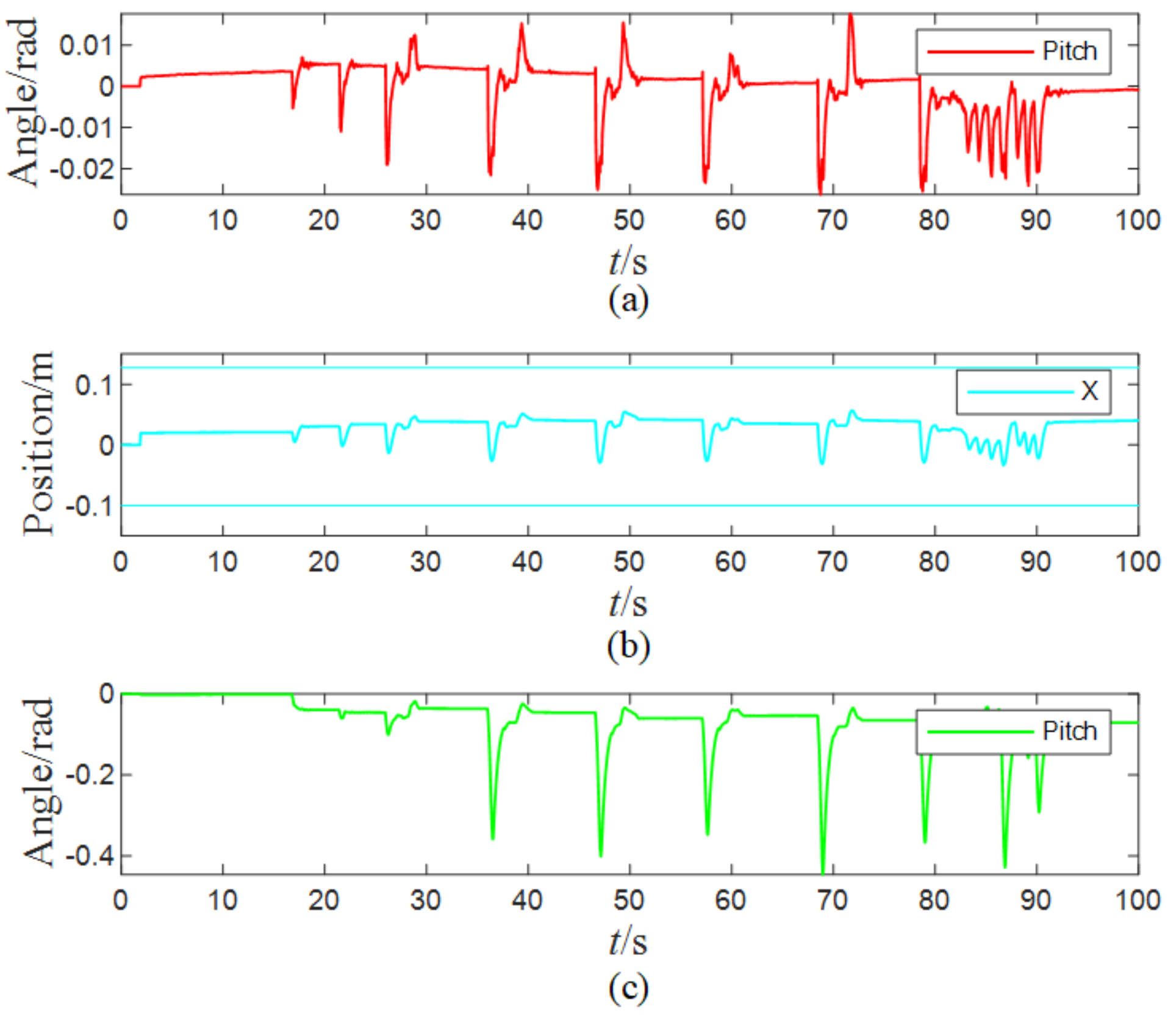}
    \caption{The measured pitch angle of right foot (a), CoM position along X-axis (b) and pitch angle of torso (c).}
    \label{fig_push_recovery_task}
\end{figure}

Several key features of the tasks are drawn in Fig. \ref{fig_push_recovery_task}. It can be seen that the peak of the foot pitch angle gradually increases as the impulses become larger. The foot pitch angle reaches up to 1.5$^\circ$ when the impulse comes to its maximum 12Ns. Theoretically, the foot pitch angle should be zero because the CoP constraint has been considered in the hierarchical optimization. Given that there is a carpet between the foot and ground, such a small pitch angle is acceptable. 

Fig. \ref{fig_push_recovery_task}(b) draws the CoM position along X-axis. The CoM position fluctuates between -29mm~56mm under the continuous impulses. Fig. \ref{fig_push_recovery_task}(c) draws the pitch angle of torso. The robot tries to rotate the upper body to preserve the foot posture and the CoM position, which looks similar to that of a human rotating its trunk to maintain balance. The maximum value of the torso pitch angle is 25.7$^\circ$. Continuing increasing the impulse will cause a balance failure due to the torso pitch angle exceeding the joint angle limits. 

For these three tasks in Fig. \ref{fig_push_recovery_task}, there always exists a tiny stability error although the tasks' feedback control law works. Such a control command resulted from the tiny error will not drive the joint to move. In the video, the robot behaves that it can not recover to its original state after the disturbances.

\begin{figure}[htp]
    \centering
    \includegraphics[width=3.5in]{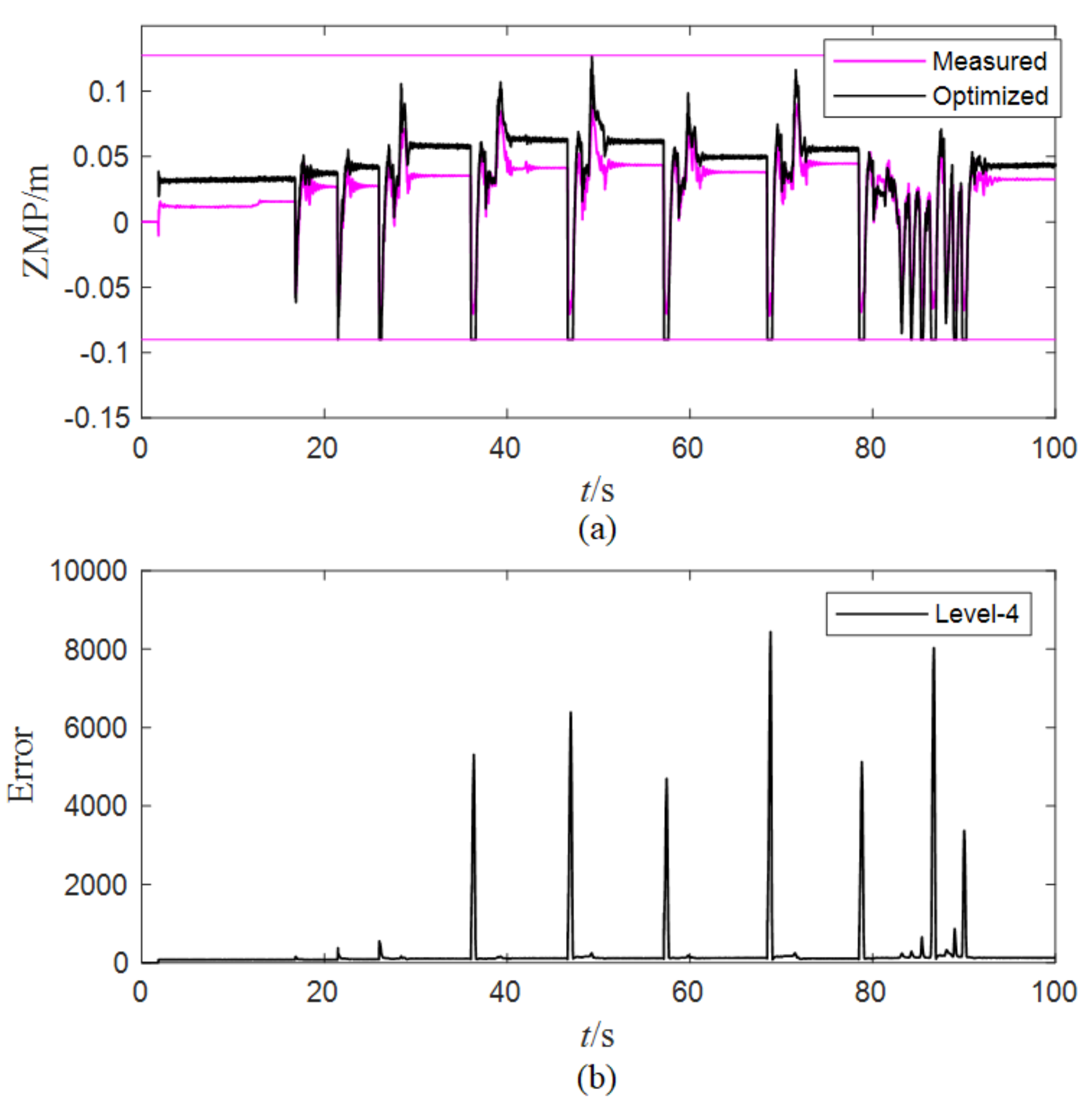}
    \caption{The measured and optimized ZMP of right foot (a) and the task error of the lowest priority task (b).}
    \label{fig_push_recovery_error}
\end{figure}

Fig. \ref{fig_push_recovery_error}(a) compares the measured ZMP with the optimized one. The measured ZMP is calculated using the force sensor while the optimized ZMP is calculated using the optimized foot contact force. Ideally, the measured ZMP should follow the optimized ZMP closely, but actually there is a slight difference. It means that all motors in the robot can not generate the required optimized foot contact force. The main reason accounting for that is the joint torque error due to the limited identification accuracy of the joint friction model in Sec. \ref{sec:exp:dynamic_parameters_identification}.

In addition, there are several times that the optimized ZMP reaches up to its boundary. It indicates a strong linear relationship between the optimal foot contact force, which directly leads to the dimensionality reduction of optimal variables. As a final result, the robot tends to sacrifice the lowest priority task due to a lack of DoFs. As proof, Fig. \ref{fig_push_recovery_error}(b) plots the task error: $ error = {\left\| {Ax - b} \right\|^2} $ of the lowest priority task. It can be seen that the task error is small enough when the first three impulses act on the robot. However, there will always be a sharp peak with the magnitude of $ {10^3} $ as long as the impulses increase to 12Ns. The huge task error will deteriorate the control performance. The robot is originally designed to show its torso compliance accodrding to the PD parameters, but the compliance characteristic can not be ensured owning to the task error. The scarified compliance will enlarge the amplitude of the torso pitch angle, which further limits the push recovery performance.

\subsection{Balancing on a seesaw}

\begin{figure*}[htp]
    \centering
    \includegraphics[width=6.0in]{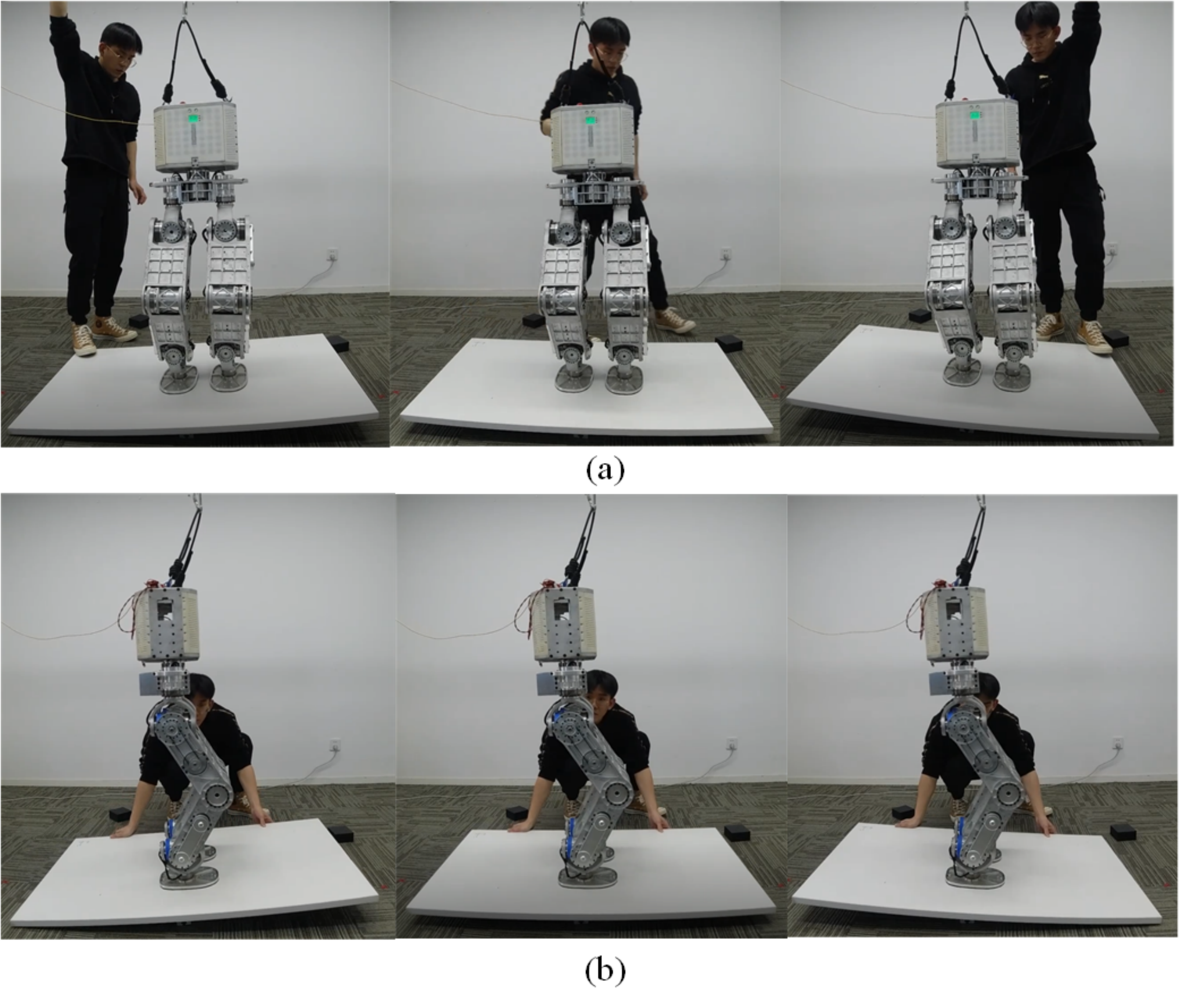}
    \caption{The balancing behaviors when the seesaw rotates along X-axis (a), and Y-axis (b).}
    \label{fig_seesaw}
\end{figure*}

\begin{figure}[htp]
    \centering
    \includegraphics[width=3.5in]{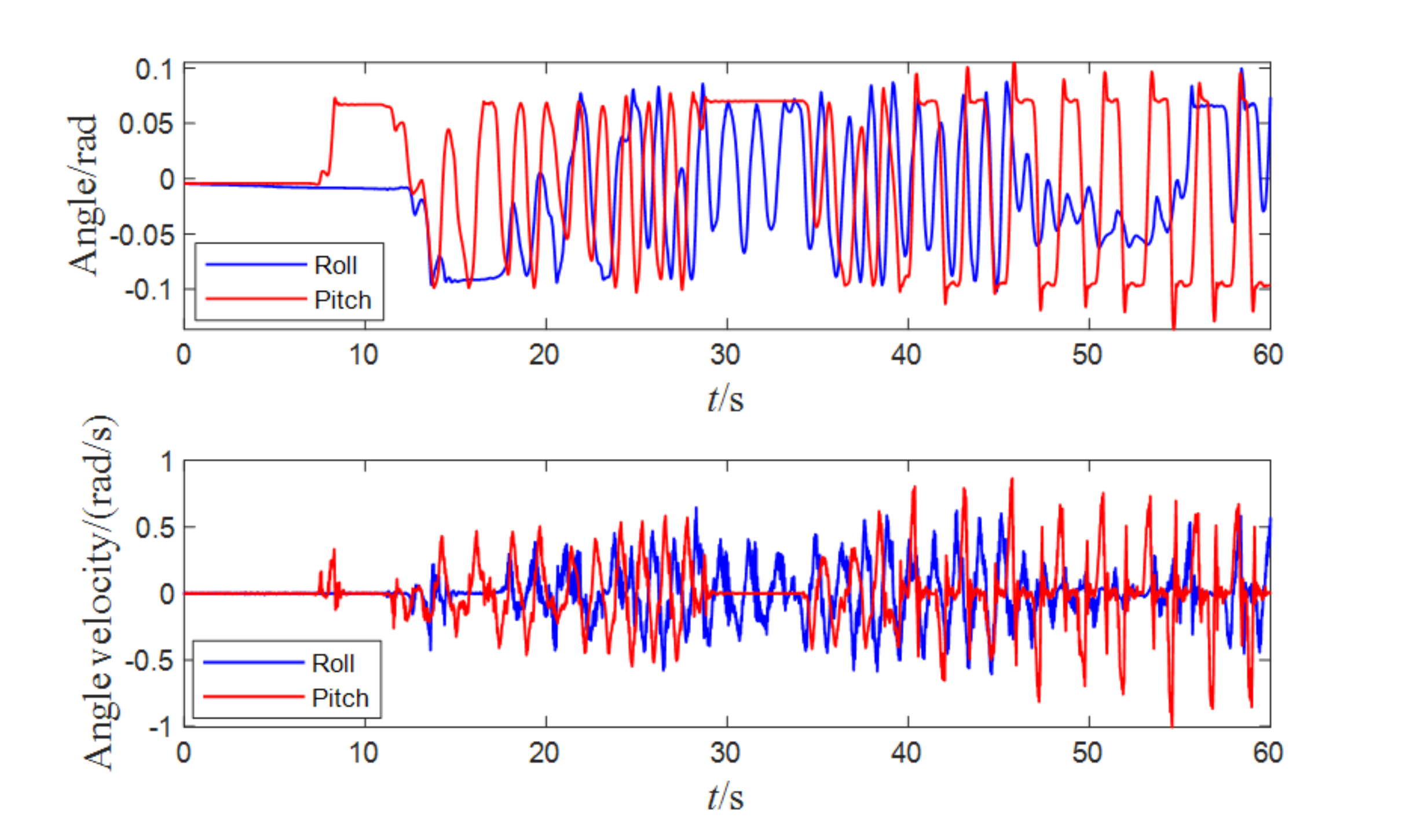}
    \caption{The measured orientation and angular velocity of right foot.}
    \label{fig_seesaw_state}
\end{figure}

The robot balances on a seesaw with the inclination disturbances along X-axis and Y-axis. Fig. \ref{fig_seesaw} shows how the robot adapts to the inclined surface to keep balance, and Fig. \ref{fig_seesaw_state} plots the measured orientation and angular velocity of right foot. Unfortunately, no additional IMU is mounted on the seesaw to measure its real movement. Here, the measured state of the foot is approximated as the estimation of seesaw because the ZMP resides inside the foot polygon throughout the test.

The amplitude of inclined angles along both axes reach up to 6$^\circ $, as shown in Fig. \ref{fig_seesaw_state}. Meanwhile, the maximum angular velocity along Y-axis is 1rad/s which is slightly larger than that along X-axis (0.6rad/s). The reason mainly relies on that the joint friction has a minor influence on the balance performances when the seesaw inclines along Y-axis. The robot merely needs to modulate its ankle pitch joint to adapt to the inclined seesaw along Y-axis while needs to modulate its ankle roll joint as well as the length of both legs to adapt to the inclined seesaw along X-axis. Here, the presented angular velocity data has been processed by a low-pass filter with a cutoff frequency of 20Hz. 

\begin{figure}[htp]
    \centering
    \includegraphics[width=3.5in]{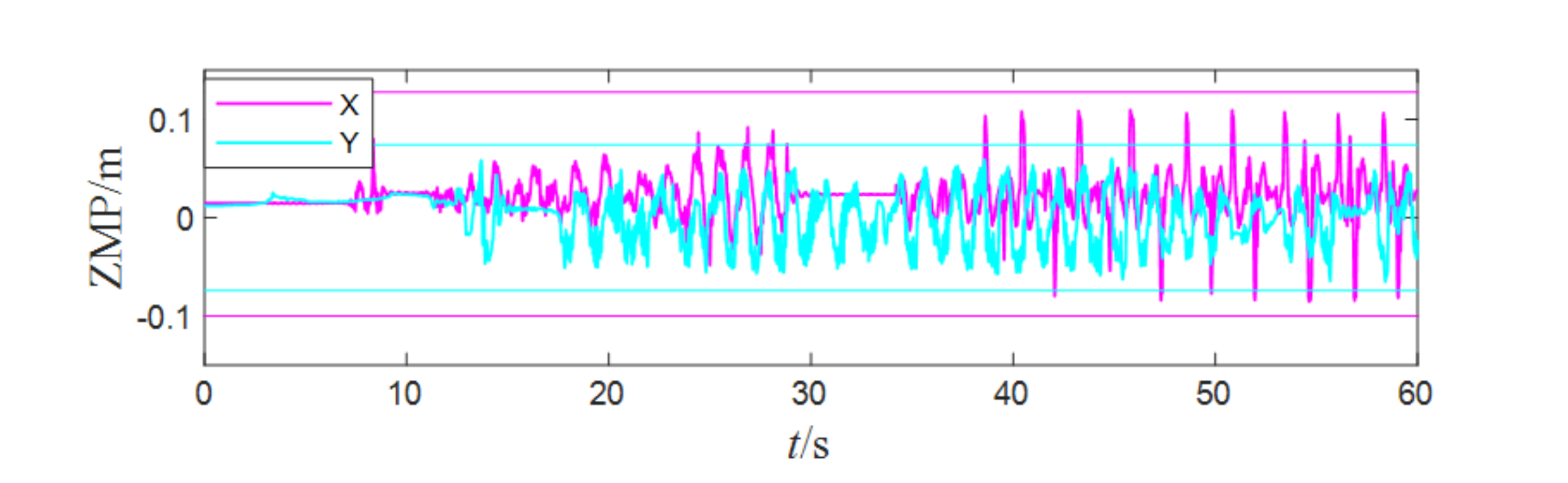}
    \caption{The measured ZMP of right foot.}
    \label{fig_seesaw_zmp}
\end{figure}

Fig. \ref{fig_seesaw_zmp} shows the response of right foot's ZMP during the disturbances. The components of ZMP along X and Y axes did not exceed the constrained boundary defined by foot geometry. All these proves that the task planner works well and the robot can adjust the tasks' target to resist the disturbance from the seesaw.

\subsection{Push recovery on two moving skateboards}

\begin{figure*}[htp]
    \centering
    \includegraphics[width=6.0in]{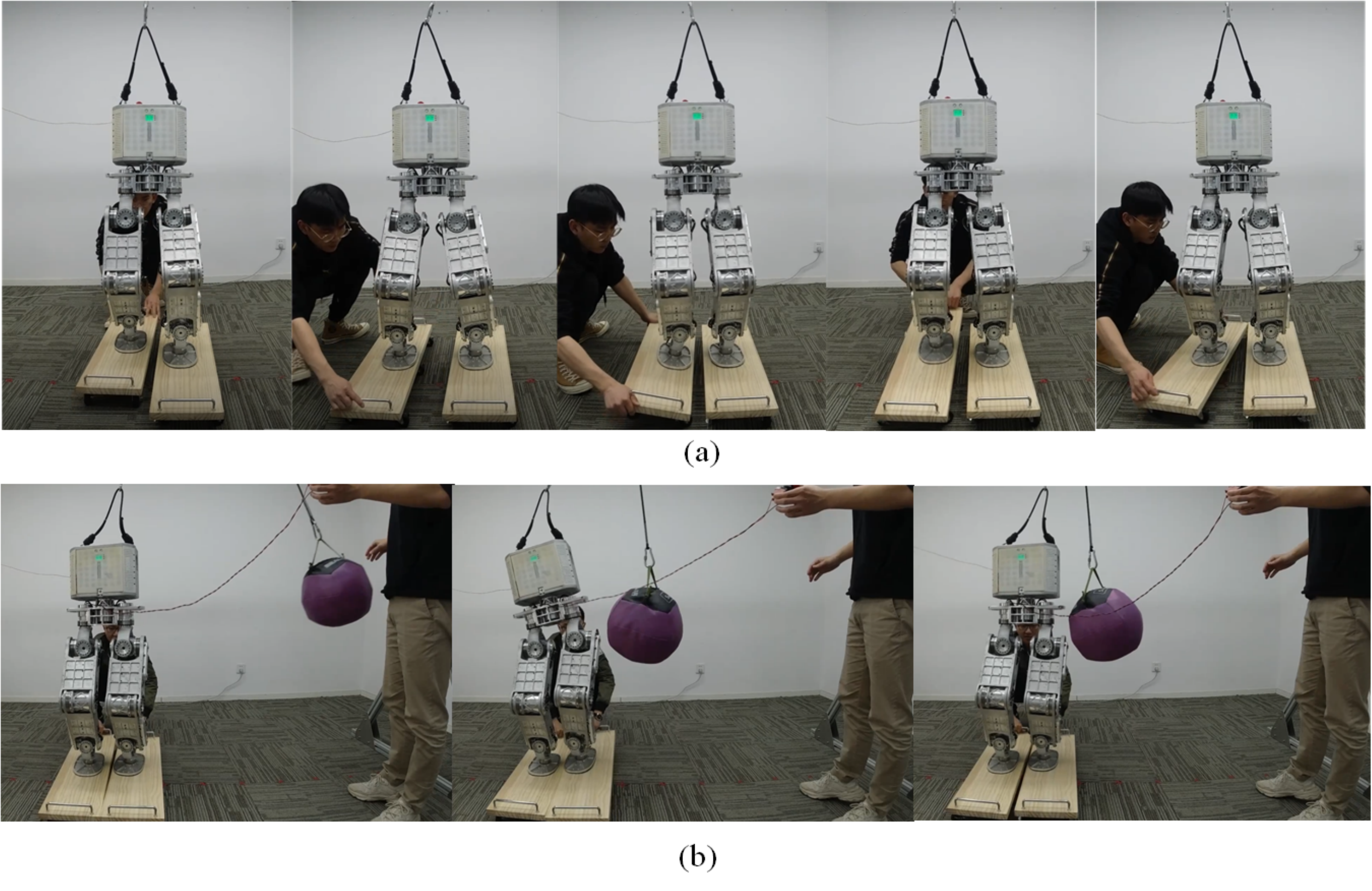}
    \caption{The balancing behaviors when the right support moves in all directions (a), and the balancing behaviors in push recovery on moving support scenario (b).}
    \label{fig_moving_support}
\end{figure*}

\begin{figure}[htp]
    \centering
    \includegraphics[width=3.5in]{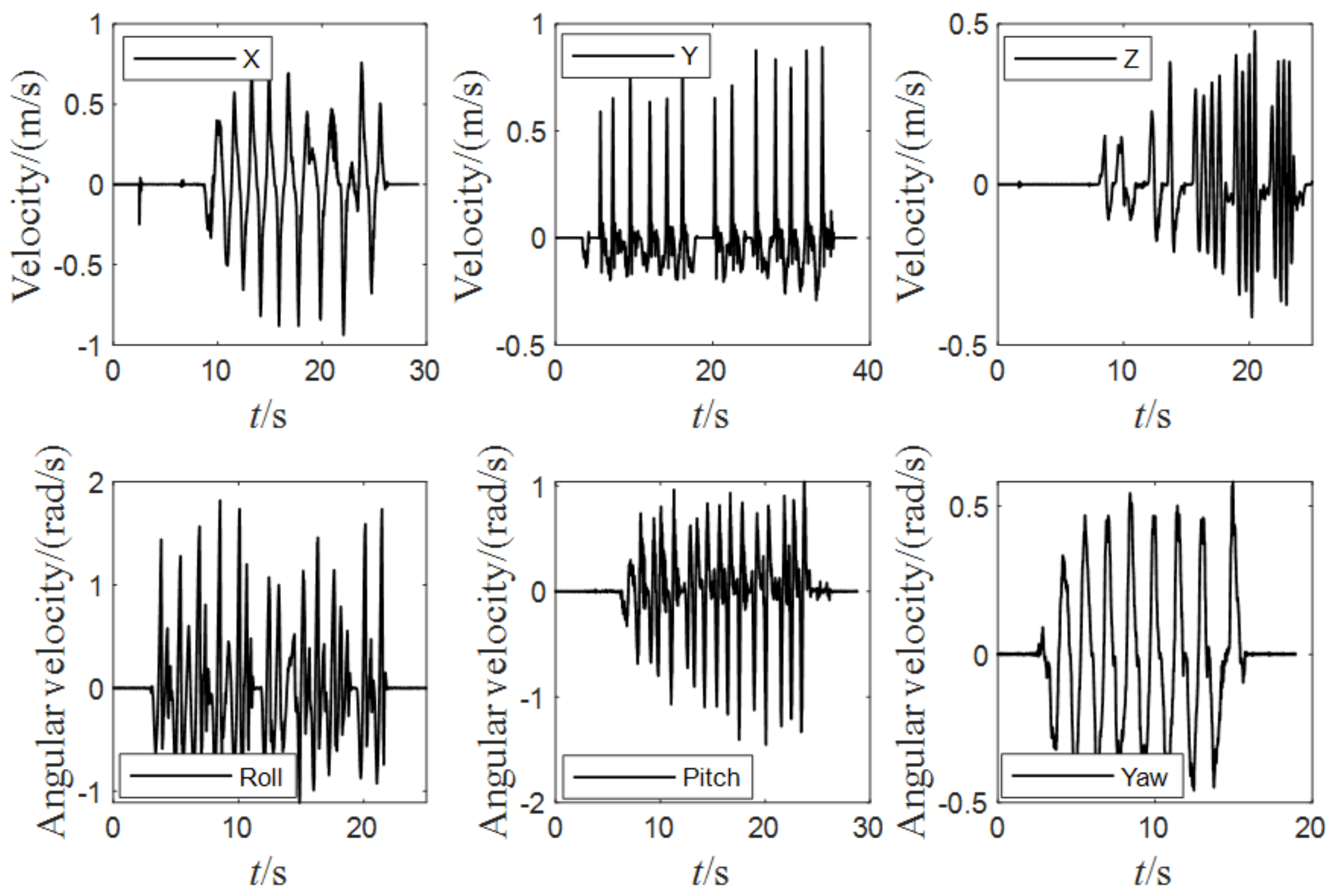}
    \caption{The measured velocity of right foot with each direction tested separately.}
    \label{fig_moving_support_velocity}
\end{figure}

The final experiment is balance maintenance on two moving skateboards as shown in Fig. \ref{fig_moving_support}. The two feet of the robot rest on two moving skateboards separately and suffer inclination and shift disturbances independently. To evaluate the anti-disturbance capability, the right moving skateboard is actuated by hands to translate along X, Y-axis and rotate along X, Y and Z-axis while the left one is locked (Fig. \ref{fig_moving_support}(a)). Fig. \ref{fig_moving_support_velocity} plots the measured 
velocity of the right foot with each direction tested separately. The shift disturbance along Z-axis is measured indirectly by rotating the skateboard about Y-axis where the foot will rise as the tilt angle increases. The robot can resist the moving skateboard disturbance with the maximum velocities 0.94m/s, 0.89m/s, 0.47m/s along X, Y, Z-axis and maximum angular velocities 1.8rad/s, 1.4rad/s, 0.5rad/s along X, Y, Z-axis. 

The robot can also maintain good balance when the two skateboards not only have different inclination angles but translate back and forth with out of phase velocity. Meanwhile, when 8Ns impulses along Y-axis are exerted on the robot as the skateboards keep moving, the robot generates a large torso rotation to keep balance (Fig. \ref{fig_moving_support}(b)).

\begin{figure}[htp]
    \centering
    \includegraphics[width=3.5in]{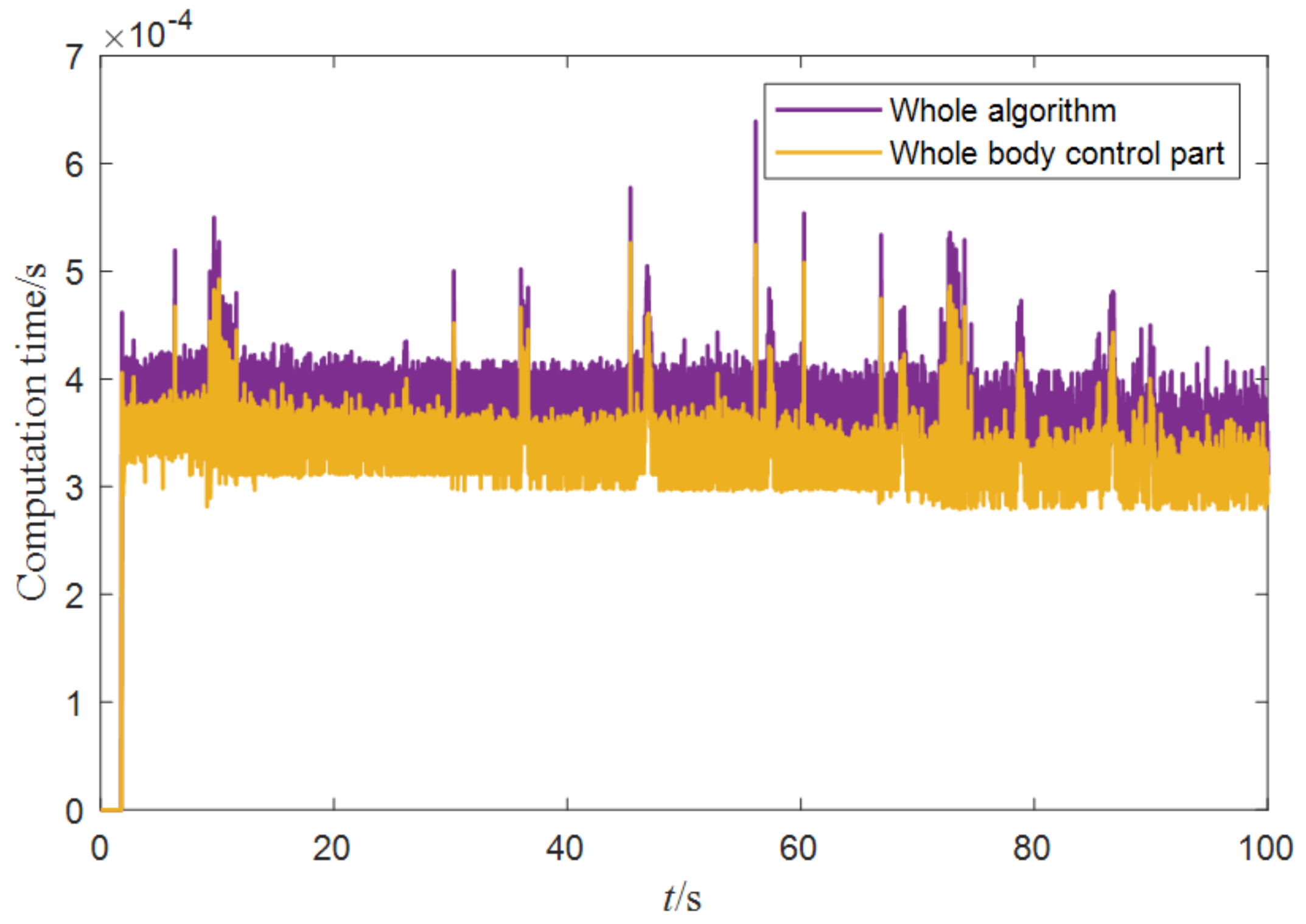}
    \caption{The computation time of the algorithm.}
    \label{fig_moving_support_time}
\end{figure}

To achieve a 1KHz control loop, the on-board computer needs to solve the hierarchical optimization which contains four quadratic optimization problems in 1ms. Fig. \ref{fig_moving_support_time} plots the computation time of the whole algorithm including the state estimation, trajectory planning, whole-body control and joint-level control parts. The average and the maximum computation time is 0.363ms and 0.639ms. As the most time-consuming portion, the computation time of the whole-body control part is also plotted here in yellow line. The average computation time is 0.321ms and it takes up about 88 percents of the whole time, leaving 0.042ms for the other parts. In the whole-body control part, we need to solve the quadratic optimization and the null space projection matrix sequentially. About 72\% of the computation time is used for quadratic optimization and the left 28\% is used for the null space projection matrix. The quadratic optimization is solved through a C++ open source QP solver, qpOASES \cite{2014ferreau}, which implements the active set algorithm. The null space projection matrix needs to calculate the pseudo inverse of the matrix first and the complete orthogonal decomposition algorithm implemented in the Eigen matrix library is used here.

\section{Conclusions}

The purpose of this paper is endowing the proprioceptive actuated humanoid robot with the ability of dynamic balance. To that end, the tasks and constraints are assigned and customized with a reasonable hierarchy. The hierarchical whole-body control is reduced for real-time computation and implemented via a computationally efficient WBC software. A modular master control system UBTMaster characterized by the real-time communication and powerful computing capability is also designed. With the aid of these software and hardware, users can easily develop their real-time applications on their robots without rewriting the WBC solver code. For the robot with proprioceptive actuation, key dynamic parameters are identified to cover the nonlinear joint friction and model inaccuracy problems. The identified model are accurate enough so that the predicted torque can follows the measured one closely with a mean residual less than 1Ns.

With these three aspects being well considered, the balance performance of a humanoid robot Walker3 is fully tested in various scenarios. The robot can maintain balance under continuous impulses along X and Y axes with the maximum value up to 12Ns. It tends to rotate its upper body to preserve the foot posture and the CoM position, which is similar to the human behavior. In addition, we also find the reason that limits the push recovery performance. The dimensionality reduction in the optimal variables as they reach to the constrained boundary will lead to the the sacrifice of the lowest priority task. The higher priority tasks are not affected due to the strict hierarchy. An effective solution to the problem is providing more redundant DoFs such as adding dual-arm to the robot.

When Walker3 stands on a seesaw, it can actively adapts to the inclined surface to keep balance. Different from \cite{2016herzog}, the state of seesaw is estimated without additional IMU and then used to update the trajectory of tasks. The experimental results show that the inclined angles along both axes reach up to 6$^\circ $. Meanwhile, the maximum angular velocities along X and Y axes are 0.6rad/s and 1rad/s separately, which is about 1.7 to 2.8 times of the performance on a torque-controlled robot COMAN \cite{2013li}.

To further exploit the robot's adaptability to the uncertain disturbance, we place Walker3 on two moving skateboards and impose inclination and shift disturbances simultaneously. The robot can resist the disturbance with the maximum velocities 0.94m/s, 0.89m/s, 0.47m/s along X, Y, Z-axis and maximum angular velocities 1.8rad/s, 1.4rad/s, 0.5rad/s along X, Y, Z-axis. The robot can even resists 8Ns impulses as the two skateboards not only have different inclination angles but translate back and forth with out of phase velocity. 

All these results prove that with the strict hierarchy, real-time computation and joint friction being handled carefully, the robot with proprioceptive actuation will show an excellent balance performance which is comparable to that of a torque-controlled robot. For future work, we would like to improve the balance framework with a higher level planner to handle larger disturbances. The authors believe that combining the offline nonlinear optimization and the online model predictive control techniques will improve the robustness of the robot significantly.


%



\section*{Acknowledgment}
The authors would like to thank the anonymous reviewers for their detailed and pertinent comments.

\ifCLASSOPTIONcaptionsoff
  \newpage
\fi




\bibliographystyle{IEEEtran}
\end{document}